\documentclass[letterpaper, 10 pt, conference]{ieeeconf}  

\IEEEoverridecommandlockouts                              

\usepackage{graphics} 
\usepackage{epsfig} 
\usepackage{amsmath} 
\usepackage{amssymb}  
\usepackage{orcidlink}
\usepackage{xcolor}
\usepackage{multirow}
\usepackage{multicol}
\usepackage{verbatim}
\usepackage{algorithm}
\usepackage{algorithmic}

\newcommand{\eg}{\textit{e.g.,}}
\newcommand{\vrlpap}{\textsc{vrl}-\textsc{pap}}
\newcommand{\ganav}{\textsc{gan}av}
\newcommand{\rca}{\textsc{rca}}

\newcommand{\sterling}{\textsc{sterling}}

\let\labelindent\relax  
\usepackage{enumitem}

\newcommand{\patern}{\textsc{patern}}

\usepackage[normalem]{ulem}

\definecolor{green}{RGB}{11,155,13}

\title{\LARGE \bf
Wait, That Feels Familiar: Learning to Extrapolate Human Preferences for
Preference-Aligned Path Planning
}

\author{Haresh Karnan$^{1*}$, Elvin Yang$^{2*\dagger}$, Garrett Warnell$^{3,4}$, Joydeep Biswas$^{3}$, Peter Stone$^{3,5}$ 
\thanks{$^{1}$Walker Department of Mechanical Engineering, The University of Texas at Austin,
        {\tt\small haresh.miriyala@utexas.edu}}%
\thanks{$^{2}$University of Michigan, Ann Arbor,
        {\tt\small eyy@umich.edu}}%
\thanks{$^{3}$Department of Computer Science, The University of Texas at Austin,
        {\tt\small \{joydeepb, pstone\}@cs.utexas.edu}}%
\thanks{$^{4}$Army Research Laboratory,
        {\ttfamily garrett.a.warnell.civ@army.mil}}%
\thanks{$^{5}$Sony AI, North America}%
\thanks{*Equal Contribution, sorted alphabetically by last name}
\thanks{$\dagger$ Work done while at UT Austin}
}

\begin{document}

\maketitle
\thispagestyle{empty}
\pagestyle{empty}

\begin{abstract}

Autonomous mobility tasks such as last-mile delivery require reasoning about operator-indicated preferences over terrains on which the robot should navigate to ensure both robot safety and mission success. However, coping with \emph{out of distribution} data from novel terrains  or appearance changes due to lighting variations remains a fundamental problem in visual terrain-adaptive navigation.
Existing solutions either require labor-intensive manual data re-collection and labeling or use hand-coded reward functions that may not align with operator preferences.
In this work, we posit that operator preferences for visually novel terrains, which the robot should adhere to, can often be extrapolated from established terrain preferences within the \emph{inertial-proprioceptive-tactile} domain.
Leveraging this insight, we introduce \textit{Preference extrApolation for Terrain-awarE Robot Navigation} (\patern{}), a novel framework for extrapolating operator terrain preferences for visual navigation.
\patern{} learns to map inertial-proprioceptive-tactile measurements from the robot's observations to a  representation space and performs nearest-neighbor search in this space to estimate operator preferences over novel terrains.
Through physical robot experiments in outdoor environments, we assess \patern{}'s capability to extrapolate preferences and generalize to novel terrains and challenging lighting conditions. Compared to baseline approaches, our findings indicate that \patern{} robustly generalizes to diverse terrains and varied lighting conditions, while navigating in a preference-aligned manner.
\end{abstract}

\section{INTRODUCTION}
To ensure the safety, mission success, and efficiency of autonomous mobile robots in outdoor settings, the ability to visually discern distinct terrain features is paramount. This necessity stems not only from direct implications for robot functionality but also from the operator-indicated terrain preferences that the robot must adhere to. Often, these preferences are motivated by the desire the protect delicate landscapes, such as flower beds, or to mitigate potential wear and tear on the robot by avoiding hazardous surfaces. However, during autonomous operations, ground robots frequently face unfamiliar terrains~\cite{viikd, marcoanymal} and dynamic real-world conditions such as varied lighting, that lie outside the distribution of visually recognized terrains where operator preferences have been pre-defined. This mismatch presents significant challenges for vision-based outdoor navigation~\cite{xuesusurvey}.

\begin{figure}
        \centering
        \includegraphics[width=0.475\textwidth]{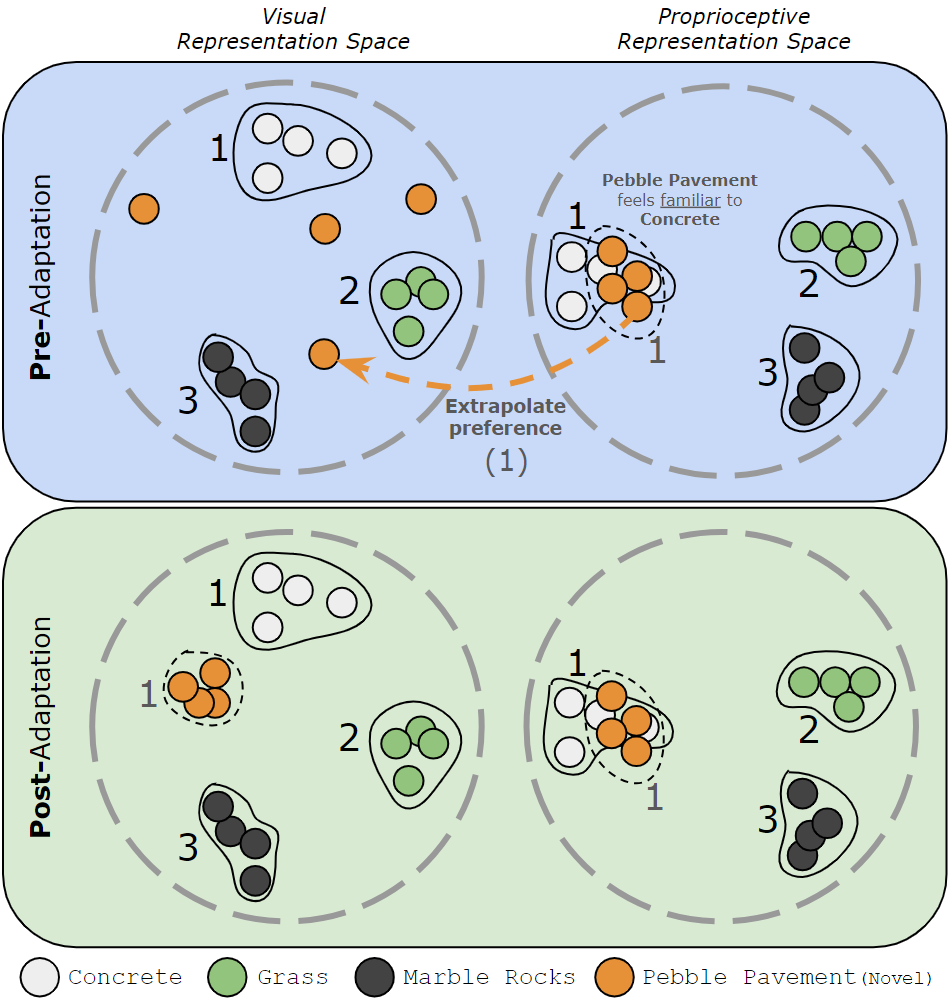}
        \caption{An illustration of the intuition behind preference extrapolation in \patern{}. Operator preferences of the three known terrains are marked numerically, with 1 being the most preferred and 3 being the least preferred. In the pre-adaptation stage, a novel terrain ($\texttt{pebble pavement}$) is encountered and the preference order of its nearest neighbor ($\texttt{concrete}$) inferred from proprioceptive representations is transferred (extrapolated) to the corresponding samples in the visual representation space. The extrapolated preference order is used to update both the visual representations and the visual preference function. The post-adaptation stage shows extrapolated preferences in the updated visual representation space for the novel terrain.}
        \label{fig:patern_intuition}
\end{figure}

Equipping robots with the capability to handle novel terrain conditions for preference-aligned path planning is a challenging problem in visual navigation. Prior approaches to address this problem include collecting more expert demonstrations \cite{bojarskie2e, yanne2eoffroad, vrlpap}, labeling additional data \cite{ganav, rugd, jiang2020rellis3ddataset}, and utilizing hand-coded reward functions to assign traversability costs \cite{rca, badgr, terrapn}. While these approaches have been successful at visual navigation, collecting more expert demonstration data and labeling may be labor-intensive and expensive, and utilizing hand-coded reward functions may not always align with operator preferences. We posit that in certain cases, while the terrain may look visually distinct in comparison to prior experience, similarities in the inertial-proprioceptive-tactile space may be leveraged to extrapolate operator preferences over such terrains, that the robot must adhere to. For instance, assuming a robot has experienced $\texttt{concrete pavement}$ and $\texttt{marble rocks}$, and prefers the former over the latter (as expressed by the operator), when the robot experiences a visually novel terrain such as $\texttt{pebble pavement}$ which feels inertially similar to traversing over $\texttt{concrete pavement}$, it is more likely that the operator might also prefer $\texttt{pebble pavement}$ over $\texttt{marble rocks}$. While it is not possible to know the operator's true preferences without querying them, we submit that in cases where the operator is unavailable, hypothesizing preferences through extrapolation from the inertial-proprioceptive-tactile space is a plausible way to estimate traversability preferences for novel terrains.

Leveraging the intuition of extrapolating operator preferences for visually distinct terrains that are familiar in the inertial-proprioceptive-tactile space (collectively known as \textit{proprioceptive} for brevity), we introduce \textit{Preference extrApolation for Terrain-awarE Robot Navigation} (\patern{}) \footnote{A preliminary version of this work was presented at the PT4R workshop at ICRA 2023 \cite{paternicra}}, a novel framework for extrapolating operator terrain preferences for visual navigation. \patern{} learns a proprioceptive latent representation space from the robot's prior experience and uses nearest-neighbor search in this space to estimate operator preferences for visually novel terrains. Fig. \ref{fig:patern_intuition} provides an illustration of the intuition behind preference extrapolation in \patern{}. We conduct extensive physical robot experiments on the task of preference-aligned off-road navigation, evaluating \patern{} against state-of-the-art approaches, and find that \patern{} is empirically successful with respect to preference alignment and in adapting to novel terrains and lighting conditions seen in the real world.

\section{RELATED WORK}
In this section, we review related work in visual off-road navigation, with a focus on preference-aligned path planning.

\subsection{Supervised Methods}
To learn terrain-aware navigation behaviors, several prior methods have been proposed that used supervised learning from large curated datasets \cite{rugd, jiang2020rellis3ddataset} as supervision to pixel-wise segment terrains \cite{ganav}. Guan et al. \cite{ganav} propose a transformer-based architecture (\ganav{}) to segment terrains, and manually assign traversability costs for planning. While successful at preference-aligned navigation, fully-supervised methods suffer from domain shift on novel terrains and may require additional labeling.

\subsection{Self-Supervised Methods}
To alleviate the need for large-scale datasets for visual navigation, several self-supervised learning methods have been proposed that learn from data collected on the robot \cite{frey2023fast}. Specifically, prior methods in this category have explored using inertial Fourier features \cite{rca}, contact vibrations \cite{brooks2007self}, proprioceptive feedback \cite{loquercio2022learning}, odometry errors \cite{terrapn}, future predictive models \cite{badgr}, acoustic features \cite{ser}, and trajectory features \cite{wher2walk} to learn traversability costs for visual navigation. While successful in several visual navigation tasks such as \textit{comfort-aware navigation} \cite{rca}, such methods use a hand-coded reward/cost model to solve a specific task and do not reason about operator preferences over terrains. In contrast with prior methods, \patern{} utilizes the prior experience of the robot and extrapolates operator preferences to novel terrains.

Sikand et al. propose \vrlpap{} \cite{vrlpap} in which both a visual representation and a visual preference cost are learned for preference-aligned navigation. Similarly, \sterling \cite{sterling} introduces a self-supervised representation learning approach for visual representation learning. However, a limitation for both \vrlpap{} and \sterling{} is their dependence on additional human feedback when dealing with novel terrains — feedback that might not consistently be available during deployment. Distinct from \vrlpap{} and \sterling{}, \patern{} focuses on extrapolating operator preferences from known terrains to visually novel terrains.

\section{PRELIMINARIES}
We formulate preference-aligned planning as a local path-planning problem in a state space $\mathcal{S}$, with an associated action space
$\mathcal{A}$. The forward kino-dynamic transition function is denoted as $\mathcal{T}:
\mathcal{S} \times \mathcal{A} \rightarrow \mathcal{S}$ and we assume
that the robot has a reasonable model of $\mathcal{T}$ (\eg{} using parametric system identification~\cite{seegmiller2013vehicle} or a learned kino-dynamic
model~\cite{xuesuikd,viikd,optimfkd}), and that the robot can execute actions
in $\mathcal{A}$ with reasonable precision. 
For ground vehicles, a common choice for $\mathcal{S}$ is $\mathrm{SE}(2)$, which represents the robot's x and y position on the ground plane, as well as its orientation $\theta$. 


The objective of the path-planning problem can be expressed as finding the optimal trajectory
$\Gamma^* = \underset{\Gamma}{\arg \min}\ J(\Gamma, G)$ to the goal $G$, using any planner (e.g. a sampling-based motion planner like \textsc{dwa} \cite{dwa}) while minimizing an objective function $J(\Gamma, G)$, $J : (\mathcal{S}^N, \mathcal{S}) \rightarrow \mathbb{R}^+$. Here, $\Gamma = \{s_1, s_2, \dots, s_N\}$ denotes a sequence of states. The sequence of states in the optimal trajectory $\Gamma^*$ is then translated into a sequence of actions, using a 1-D time-optimal controller, to be played on the robot. For operator preference-aligned planning, the objective function $J$ is articulated as,
\begin{equation}
    J(\Gamma, G) = J_G(\Gamma(N), G) + J_P(\Gamma),
\end{equation}
\noindent

Here, $J_G$ denotes cost based on proximity of the robot's state to the goal $G$, while $J_P$ imparts a cost based on terrain preference. 
Crucially, $J_P$ is designed to capture operator preferences over different terrains; less preferred terrains incur a higher cost. Though earlier studies leverage human feedback to ascertain $J_P$ for unfamiliar terrains \cite{vrlpap, sterling}, in this work, we hypothesize that in certain situations, operator preferences for novel terrains can be extrapolated from known terrains, obviating operator dependency during real-world deployment. Thus, our novel contribution is a self-supervised framework for extrapolating $J_P$ from known terrains to visually novel terrains by leveraging inertial-proprioceptive-tactile observations of a robot, without inquiring additional human feedback.




\section{APPROACH}

In this section, we present \textit{Preference extrApolation for Terrain-awarE Robot Navigation} (\patern{}), a novel framework for extrapolating operator preferences for preference-aligned navigation. We first detail an existing framework for terrain-preference-aligned visual navigation. We then introduce \patern{} for self-supervised extrapolation of operator preferences from known terrains to visually novel terrains by leveraging proprioceptive feedback.

\subsection{A Two-Step Framework for Preference-Aligned Planning}
\label{subsec:two_step_framework}
For real-time preference-aligned planning, inspired from earlier studies \cite{vrlpap, sterling}, we postulate that $J_P(\Gamma)$ can be estimated in a two-step approach from visual observations of patches of terrain at $s \in \mathcal{S}$ along $\Gamma$. Let $O \in \mathcal{O}$ represent these observations. We denote $\Pi$ as a projection operator that extracts visual observation $O$ of terrain at $s$ by yielding image patches from homography-transformed bird's eye view images \cite{vrlpap, sterling}. First, a visual encoder, denoted as $f_{vis}$, maps $O$ from the RGB space to a latent vector $\phi_{vis} \in \Phi_{vis}$ such that observations from identical terrains cluster closely in $\Phi_{vis}$ and are distinct from those of differing terrains. Next, a real-valued preference utility is estimated from $\phi_{vis}$ using a learned preference utility function $u_{vis}: \Phi_{vis} \rightarrow \mathbb{R}^+$ trained with ranked preferences of terrains, derived either from demonstrations \cite{vrlpap}, or by active querying \cite{sterling}. Adopting the popular formulation of Zucker et al. \cite{zucker2011optimization}, we train the utility function with the margin-based ranking loss \cite{BMVC2016_119}. To estimate $J_P(\Gamma)$ during planning, we employ an exponential cost formulation given by, $J_P(\Gamma) = \sum_{s \in \Gamma} e^{-u_{vis}[f_{vis}(\Pi(s))]}$, which we find works well in practice. This two-step framework for estimating $J_P(\Gamma)$ has been utilized successfully in recent works \cite{vrlpap, sterling} for operator preference-aligned off-road navigation. Training details of the visual encoder and the utility function are provided in Section \ref{sec:implementation_details}.

While the above two-step framework effectively handles known terrains with pre-defined preferences, it faces challenges when the robot encounters visually novel terrains that lie beyond the training distribution of $f_{vis}$ and $u_{vis}$. Towards addressing this problem, the primary contribution of our work is a self-supervised framework to extrapolate operator preferences to novel terrains and adapting $f_{vis}$ and $u_{vis}$ to ensure successful preference alignment.

\subsection{Extrapolating Preferences for Visually Novel Terrains}
Leveraging the intuition that in addition to visual appearance, operator preferences over terrains are likely also based on the ``feel" of the underlying terrain such as bumpiness, stability, or traction, we posit that in many situations, operator preferences for novel terrains can be deduced by relating the proprioceptive modality to known terrains. Utilizing these rich, alternate data sources offers deeper insight into terrain properties, enabling us to extrapolate terrain preferences when direct operator feedback is unavailable. Upon initially encountering a novel terrain, before undergoing any adaptation, we designate this stage as the \textit{pre-adaptation} phase. During this phase, the visual encoder and utility function operate based on previously known operator preferences. However, once preferences are extrapolated and the visual encoder and utility functions are subsequently retrained to adapt, the system progresses to the \textit{post-adaptation} phase, as shown in Fig. \ref{fig:patern_intuition}.

\textbf{Inertial-Proprioceptive-Tactile Encoder and Utility Function:} In \patern{}, in addition to the visual encoder, we introduce a non-visual encoder that independently processes the inertial, proprioceptive (joint angles and velocities), and tactile feet data---collectively referred to as \textit{proprioception} for brevity---observed by the robot as it traverses a terrain. This encoder maps proprioception observations into a proprioceptive representation space $\Phi_{pro}$, such that representations $\phi_{pro} \in \Phi_{pro}$ of the same terrain are closely clustered whereas those of distinct terrains are farther apart. Additionally, a utility function $u_{pro}: \Phi_{pro} \rightarrow \mathbb{R}^+$ maps the proprioceptive representation vector $\phi_{pro} \in \Phi_{pro}$ to a real-valued preference utility, similar to the visual utility function.  Note that, to estimate $J_P(\Gamma)$ during deployment, we only use $u_{vis}$ and not $u_{pro}$ since we cannot observe the proprioceptive components of a future state without traversing the terrain first. 

\textbf{Pre-Adaptation Phase:} While traversing known terrains that are in-distribution, the visual and proprioceptive utility values tend to align closely. However, for visually novel terrains, discrepancies often emerge between the utility values predicted from the visual and proprioceptive modalities. In \patern{}, we utilize the mean-squared error between the predicted utilities as a signal to detect visually novel, out-of-distribution terrains. Although any novelty detection mechanism can be integrated within \patern{}, such as the uni-modal approach by Burda et al. \cite{explorernd}, our primary focus is on a framework that extrapolates operator preferences for novel terrains. Moreover, any foundational approach employing the two-step framework for preference cost estimation \cite{vrlpap, sterling, trex}, as elaborated in Subsection \ref{subsec:two_step_framework}, can be utilized in the pre-adaptation phase. For clarity, we use the notation $\patern{}^-$ to represent the baseline algorithm in its unadapted state and $\patern{}^+$ to indicate the updated model in the post-adaptation phase.

\textbf{Extrapolating Operator Preferences:} Given a novel terrain segment for which operator preferences are unknown, we propose to self-supervise preference assignment by first clustering its proprioceptive representations $\phi_{pro}$ and then associating it with the closest known existing cluster in $\Phi_{pro}$, assigning the same operator preference as the known-cluster, as illustrated in Fig. \ref{fig:patern_intuition}. Following this self-supervised preference assignment, the visual encoder and visual utility function for novel terrain segments are finetuned by aggregating newly gathered experience with existing data.

\begin{figure}
    \centering
    \includegraphics[width=0.475\textwidth]{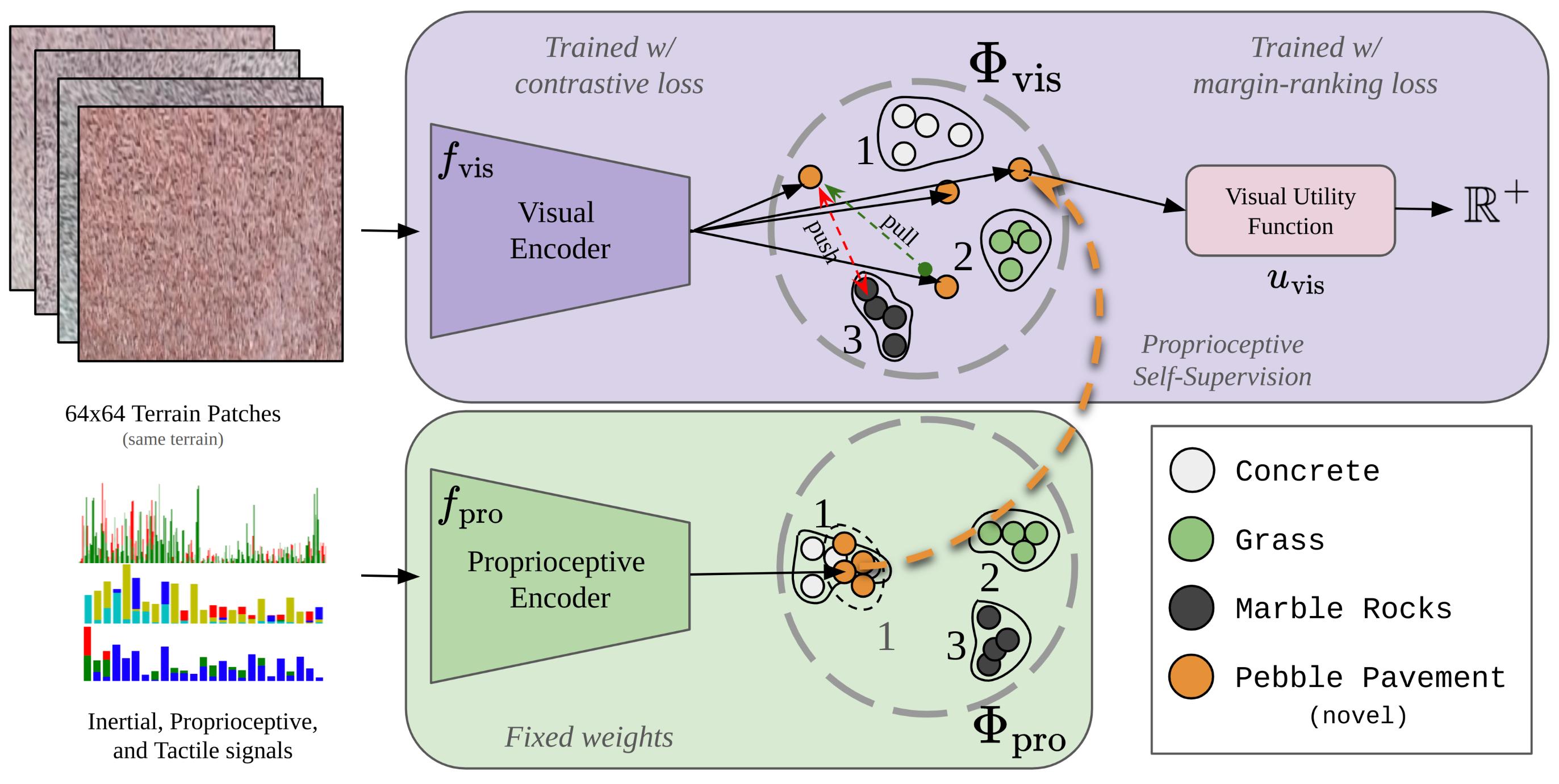}
    \caption{An illustration of the training setup for preference extrapolation proposed in \patern{}. We utilize two encoders to map visual and inertial-proprioceptive-tactile samples to $\Phi_{vis}$ and $\Phi_{pro}$ respectively. For a visually novel terrain, a preference is hypothesized and extrapolated from $\Phi_{pro}$, following which the visual encoder and utility function are retrained. 
    }
    \label{fig:training_setup}
\end{figure}


\section{Implementation Details}
\label{sec:implementation_details}


In this section, we describe the implementation details of \patern{}.  We first describe data pre-processing, followed by training details in the pre-adaptation phase. Finally, we describe adapting the visual encoder and utility function using the extrapolated preference in \patern{}.

\subsection{Data Pre-Processing}

In tandem with the visual patch extraction process used in the projection operator $\Pi$ as in prior methods \cite{vrlpap, sterling}, for every state $s_t$, we also extract a 2-second history of time-series inertial (angular velocities along the x and y-axes and linear acceleration in the z-axis), proprioceptive (joint angles and velocities), and tactile (feet depth penetration estimates) data. To ensure the resulting input data representation for training is independent of the length and phase of the signals, we compute statistical measures of center and spread as well as the power spectral density, and maintain that as the input. All the visual patches extracted with the projection operator $\Pi$ and the non-visual data for each state $s$ are then tagged with their corresponding terrain name, given that each trajectory uniquely contains a particular terrain type. In addition to processing the recorded data in the pre-adaptation phase, a human operator is queried for a full-order ranking of terrain preference labels.

\subsection{Pre-Adaptation Training}
\label{subsec:pre_adaptation_training}
We use a supervised contrastive learning formulation inspired by Sikand et al. \cite{vrlpap} to train the baseline functions $f_{vis}$ and $u_{vis}$, represented as neural networks.

\noindent \textbf{Training the Encoders:} Given labeled visual patches and proprioception data, we generate triplets for contrastive learning such that for any anchor, the positive pair is chosen from the same label and the negative pair is sourced from another label. Given such triplets, we use triplet loss \cite{tripletloss} with a margin of $1.0$ to independently train the visual and proprioception encoders through mini-batch gradient descent using the AdamW optimizer. For the visual encoder, we use a 3-layer CNN of $5\times5$ kernels, each followed by ReLU activations. This model, containing approximately 250k parameters, transforms $64\times64$ size RGB image patches into an 8-dimensional vector representation $\phi_{vis}$. Similarly, our inertial encoder is comprised of a 3-layer MLP with ReLU activations, encompassing around 4k parameters, and maps proprioceptive inputs to an 8-dimensional vector $\phi_{pro}$. To mitigate the risk of overfitting, data is partitioned in a 75-25 split for training and validation, respectively.

\noindent \textbf{Training the Utility Functions}: In our setup, the utility function is represented as a two-layer MLP with ReLU non-linearity and output activation that maps an 8-dimensional vector into a singular non-negative real value. Given ranked operator preferences of the terrains, we follow Zucker et al. \cite{zucker2011optimization} and train the visual utility function $u_{vis}$ using a margin-based ranking loss \cite{BMVC2016_119}. Furthermore, to ensure consistent predictions from $u_{vis}$ and $u_{pro}$ for both visual and non-visual observations at identical locations, we update parameters of $u_{pro}$ using the loss $\mathcal{L}_{\text{MSE}}( u_{\text{pro}}) = \frac{1}{N} \sum_{i=1}^{N} \left( sg(u_{\text{vis}}(\phi_{vis})) - u_{\text{pro}}(\phi_{pro}) \right)^2$. Here, $sg(\cdot)$ denotes the stop-gradient operation, and $\phi_{vis}$ and $\phi_{pro}$ are the terrain representations from paired visual and non-visual data, respectively, at the same location.

The functions $f_{vis}$ and $u_{vis}$ prior to adaptation are collectively termed as $\patern{}^-$, signifying their non-adapted state with respect to visually novel terrains. In our implementation, although we use supervised contrastive learning, in instances where explicit terrain labels might be absent, one can resort to self-supervised representation learning techniques, such as \sterling{} \cite{sterling}, to derive $f_{vis}$ and $u_{vis}$. \patern{} can be applied regardless of the specific representation learning approach used.



\subsection{Preference Extrapolation Training}
During deployment, if the robot encounters a visually novel terrain, both visual and inertial-proprioceptive-tactile data is recorded to be used in the adaptation phase in \patern{}, aiding in preference extrapolation and subsequent model adaptation. We refer to this collected data as the \textit{adaptation-set}. We extract paired visual and non-visual observations at identical locations from the \textit{adaptation-set} and use $f_{pro}$ to extract proprioceptive representations $\phi_{pro}$. We cluster samples of $\phi_{pro}$ and perform a nearest-neighbor search against existing terrain clusters from the pre-adaptation dataset that is within a threshold $\mu$. We set this threshold to be the same as the triplet margin value of 1.0 which we find to work well in practice. This procedure seeks a known terrain that ``feels" similar to the novel terrain which then inherits the preference of its closest match.  Following this self-supervised preference extrapolation framework, the adaptation-set is aggregated with the pre-adaptation training set, and the visual encoder $f_{vis}$ is retrained using the procedure described in \ref{subsec:pre_adaptation_training}. Additionally, the visual utility function $u_{vis}$ is retrained with the extrapolated preference for the novel terrain. The updated functions $f_{vis}$ and $u_{vis}$ are collectively referred to as $\patern{}^+$. Figure \ref{fig:training_setup} illustrates retraining and preference extrapolation as described above.

\section{EXPERIMENTS}
In this section, we describe the physical robot experiments conducted to evaluate \patern{} against other state-of-the-art visual off-road navigation algorithms. Specifically, our experiments are designed to explore the following questions: 

\begin{enumerate}[label=($Q_\arabic*$)]
    \item Is \patern{} capable of extrapolating operator preferences accurately to novel terrains?
    \item How effectively does \patern{} navigate under challenging lighting scenarios such as nighttime conditions? 
    \item How well does \patern{} perform in large-scale real-world off-road conditions?
\end{enumerate}

\begin{figure}[h!]
        \centering
        \includegraphics[width=0.485\textwidth]{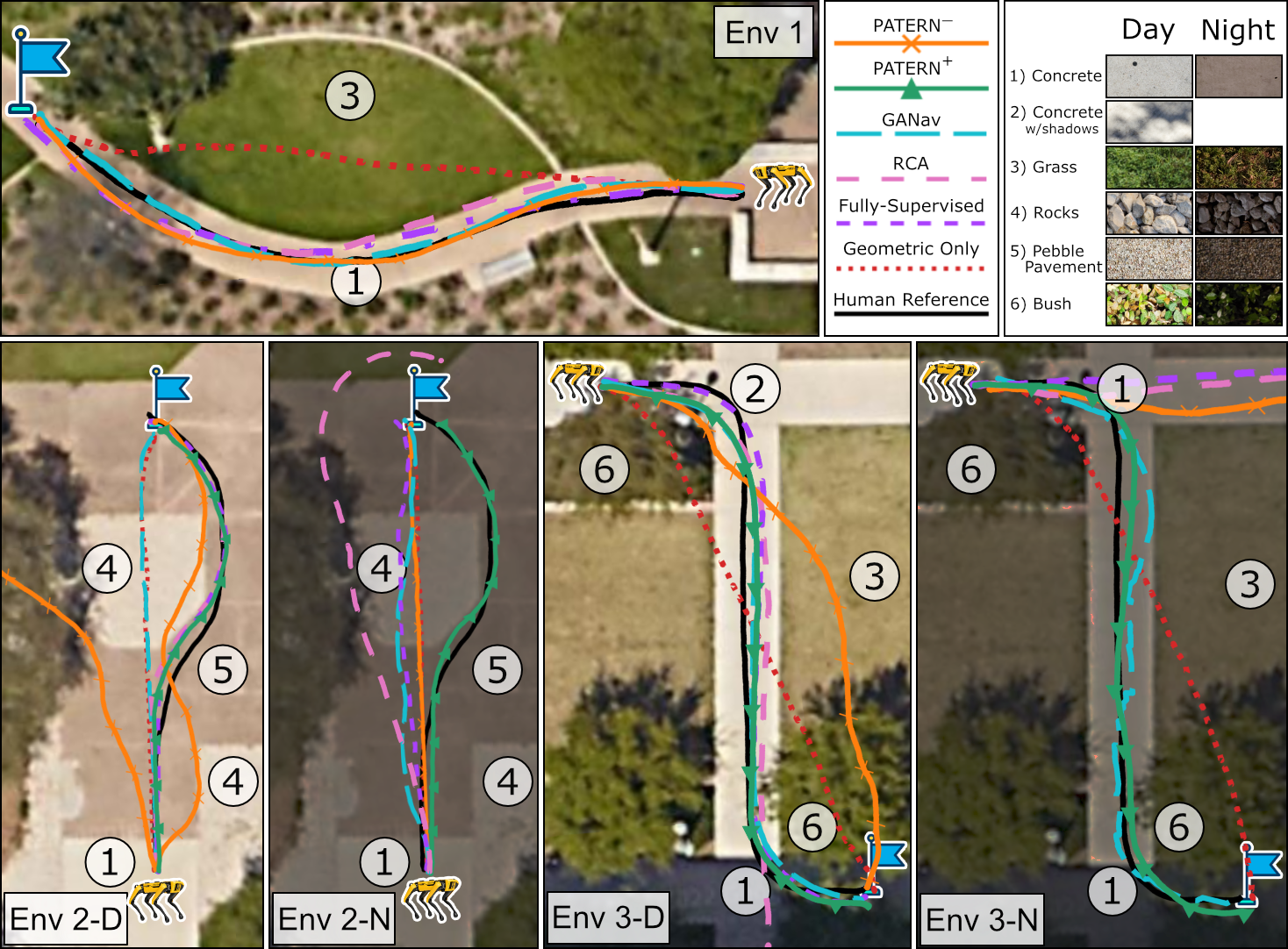}
        \caption{Trajectories traced by \patern{} and baseline approaches across three different environments and varied lighting conditions within the UT Austin campus. Note the drastic changes in the appearance of the terrain between day and night, which pose a significant challenge for visual navigation. In environments where $\patern{}^-$ fails to generalize, $\patern{}^+$ successfully extrapolates and reaches the goal in a preference-aligned manner.}
        \label{fig:robot_expts}
\end{figure}

To study $Q_1$ and $Q_2$, we execute a series of experiments consisting of short-scale outdoor navigation tasks. We then conduct a large-scale autonomous robot deployment along a 3-mile outdoor trail to qualitatively evaluate $Q_3$.
We use the legged Boston Dynamics Spot robot for all experiments, equipped with a VectorNav VN100 IMU, front-facing Kinect RGB camera, Velodyne VLP-16 LiDAR for geometric obstacle detection, and a GeForce RTX 3060 mobile GPU. For local planning, we use an open-source sampling-based motion planner called \textsc{graphnav} \cite{graphnavgithub} and augment its sample evaluation function with the inferred preference cost $L_P(\Gamma)$. For real-time planning, we run $f_{vis}$ and $u_{vis}$ on the onboard GPU.

\begin{table}[ht]
\centering
\caption{Mean Hausdorff distance relative to a human reference trajectory.}
\label{table:hausdorff}
\begin{tabular}{|l|ccccc|}
\hline

    \multicolumn{1}{|c|}{\multirow{2}{*}{\textbf{Approach}}}
    & \multicolumn{5}{c|}{\textbf{Environment}}
    \\ \cline{2-6}

    \multicolumn{1}{|c|}{}
    & \multicolumn{1}{c|}{\textbf{1}}
    & \multicolumn{1}{c|}{\textbf{2-D}}
    & \multicolumn{1}{c|}{\textbf{2-N}}
    & \multicolumn{1}{c|}{\textbf{3-D}}
    & \textbf{3-N}
    \\ \hline

    Geometric-Only
    & \multicolumn{1}{c|}{2.87}
    & \multicolumn{1}{c|}{2.34}
    & \multicolumn{1}{c|}{2.34}
    & \multicolumn{1}{c|}{3.44}
    & 3.69
    \\ \hline

    \rca{}\cite{rca}
    & \multicolumn{1}{c|}{0.84}
    & \multicolumn{1}{c|}{0.91}
    & \multicolumn{1}{c|}{6.061}
    & \multicolumn{1}{c|}{2.57}
    & 7.37
    \\ \hline

    \ganav{}\cite{ganav}
    & \multicolumn{1}{c|}{1.47}
    & \multicolumn{1}{c|}{2.98}
    & \multicolumn{1}{c|}{3.07}
    & \multicolumn{1}{c|}{0.898}
    & 1.42
    \\ \hline

    Fully-Supervised
    & \multicolumn{1}{c|}{0.58}
    & \multicolumn{1}{c|}{\textbf{0.44}}
    & \multicolumn{1}{c|}{2.735}
    & \multicolumn{1}{c|}{\textbf{0.763}}
    & 6.747
    \\ \hline \hline

    $\patern{}^-$
    & \multicolumn{1}{c|}{\textbf{0.54}}
    & \multicolumn{1}{c|}{2.31}
    & \multicolumn{1}{c|}{2.29}
    & \multicolumn{1}{c|}{2.305}
    & 5.76
    \\ \hline

    $\patern{}^+$
    & \multicolumn{1}{c|}{-}
    & \multicolumn{1}{c|}{0.56}
    & \multicolumn{1}{c|}{\textbf{1.097}}
    & \multicolumn{1}{c|}{0.86}
    & \textbf{0.763}
    \\ \hline
\end{tabular}
\end{table}

\begin{table}[ht]
\centering
\caption{Mean trajectory percentage aligned with operator preferences.}
\label{table:aligned_percent}
\begin{tabular}{|l|ccccc|}
\hline
    \multicolumn{1}{|c|}{\multirow{2}{*}{\textbf{Approach}}}
    & \multicolumn{5}{c|}{\textbf{Environment}}
    \\ \cline{2-6}

    \multicolumn{1}{|c|}{}
    & \multicolumn{1}{c|}{\textbf{1}}
    & \multicolumn{1}{c|}{\textbf{2-D}}
    & \multicolumn{1}{c|}{\textbf{2-N}}
    & \multicolumn{1}{c|}{\textbf{3-D}}
    & \textbf{3-N}
    \\ \hline

    Geometric-Only
    & \multicolumn{1}{c|}{44.0\%}
    & \multicolumn{1}{c|}{68.8\%}
    & \multicolumn{1}{c|}{68.8\%}
    & \multicolumn{1}{c|}{43.6\%}
    & 43.6\%
    \\ \hline

    \rca{}\cite{rca}
    & \multicolumn{1}{c|}{100\%}
    & \multicolumn{1}{c|}{97.3\%}
    & \multicolumn{1}{c|}{67.4\%}
    & \multicolumn{1}{c|}{100\%}
    & 99.4\%
    \\ \hline

    \ganav{}\cite{ganav}
    & \multicolumn{1}{c|}{93.9\%}
    & \multicolumn{1}{c|}{71.6\%}
    & \multicolumn{1}{c|}{71.4\%}
    & \multicolumn{1}{c|}{100\%}
    & 100\%
    \\ \hline

    Fully-Supervised
    & \multicolumn{1}{c|}{100\%}
    & \multicolumn{1}{c|}{100\%}
    & \multicolumn{1}{c|}{71.7\%}
    & \multicolumn{1}{c|}{100\%}
    & 93.6\%
    \\ \hline \hline

    $\patern{}^-$
    & \multicolumn{1}{c|}{\textbf{100\%}}
    & \multicolumn{1}{c|}{94.1\%}
    & \multicolumn{1}{c|}{71.6\%}
    & \multicolumn{1}{c|}{81.3\%}
    & 100\%
    \\ \hline

    $\patern{}^+$
    & \multicolumn{1}{c|}{-}
    & \multicolumn{1}{c|}{\textbf{100\%}}
    & \multicolumn{1}{c|}{\textbf{98.2\%}}
    & \multicolumn{1}{c|}{\textbf{100\%}}
    & \textbf{100\%}

\\ \hline
\end{tabular}
\end{table}


\subsection{Data Collection}
\label{subsec:dataprocessing}
 To collect labeled data for training $\patern{}^-$, we manually teleoperated the robot across the UT Austin campus, gathering 8 distinct trajectories each across 3 terrains: \texttt{concrete}, \texttt{grass}, and \texttt{marble rocks}. Each trajectory, five minutes long, is exclusive to a single terrain type for ease of labeling and evaluation. The pre-adaptation training data was collected in the daytime under sunny conditions. Our evaluations then centered on two preference extrapolation scenarios: one, extending to new terrains such as \texttt{pebble}-\texttt{pavement}, \texttt{concrete-with-shadows}, and \texttt{bushes}, all experienced under varying daylight conditions ranging from bright sunlight to overcast skies, and two, adapting to nighttime illumination for familiar terrains that appear visually different. In our experiments, we use the preference order $\texttt{concrete} \succ \texttt{grass} \succ \texttt{marble rocks}$.

\subsection{Quantitative Short-Scale Experiments}

We evaluate \patern{} in three environments with a variety of terrains within the UT Austin campus. We also test under two different lighting conditions, as shown in Fig. \ref{fig:robot_expts}. The primary task for evaluation is preference-aligned visual off-road navigation, in which the robot is tasked with reaching a goal, while adhering to operator preferences over terrains.

To evaluate the effectiveness of \patern{}, we compare it against five state-of-the-art baseline and reference approaches: \textbf{Geometric-Only} \cite{graphnavgithub}, a purely geometric obstacle-avoidant planner; \textbf{RCA} \cite{rca}, a self-supervised traversability estimation algorithm based on ride-comfort; \textbf{GANav}, a semantic segmentation method \footnote{\href{https://github.com/rayguan97/GANav-offroad}{https://github.com/rayguan97/GANav-offroad}} trained on RUGD dataset~\cite{rugd}; \textbf{Fully-Supervised}, an approach that utilizes a visual terrain cost function comprehensively learned using supervised costs drawn from operator preferences; and lastly, \textbf{Human Reference}, which offers a preference-aligned reference trajectory where the robot is teleoperated by a human expert. To train the \rca{} and Fully-Supervised baselines, in addition to the entirety of the pre-adaptation dataset on the 3 known terrains, we additionally collect 8 trajectories each on the novel terrains during daytime. 

In each environment, we perform five trials of each method to ensure consistency in our evaluation. For each trial, the robot is relocalized in the environment, and the same goal location $G$ is fed to the robot. In the environments where $\patern{}^-$ fails to navigate in a preference-aligned manner, we run five trials of the self-supervised $\patern{}^+$ instance that uses experience gathered in these environments to extrapolate preferences to the novel terrains or novel lighting conditions. Fig. \ref{fig:robot_expts} shows the qualitative results of trajectories traced by each method in the outdoor experiments. Only one trajectory is shown for each method unless there is significant variation between trials.
Table \ref{table:hausdorff} shows quantitative results using the mean Hausdorff distance between a human reference trajectory and evaluation trajectories of each method. Table \ref{table:aligned_percent} shows quantitative results for the mean percentage of preference-aligned distance traversed in each trajectory. Note that both the reported metrics may be high if a method does not reach the goal but stays on operator-preferred terrain. 

From the quantitative results, we see that, as expected, the $\patern{}^-$ approach is able to successfully navigate in an operator-preference-aligned manner in Env. 1, which did not contain any novel terrain types. However, $\patern{}^-$ fails to consistently reach the goal and/or navigate in alignment with operator preferences in the remaining environments.
In the daytime experiments, Env. 2 contains a novel terrain (\texttt{pebble pavement}) absent from training data for $\patern{}^-$, while Env. 3 contains both a novel terrain type (\texttt{bush}) and novel visual terrain appearances caused by tree shadows. In the nighttime experiments, all terrains contain novel visual appearances. Following deployments in Envs. 2 and 3, \patern{} extrapolates terrain preferences for new visual data using the corresponding proprioceptive data to retrain environment-specific $\patern{}^+$ instances.
In each environment that the $\patern{}^-$ model fails, the self-supervised $\patern{}^+$ model is able to successfully navigate to the respective goal in a preference-aligned manner, without requiring any additional operator feedback during deployment, addressing $Q_1$ and $Q_2$. While the fully-supervised baseline more closely resembles the human reference trajectory compared to $\patern{}^+$ during the day in Envs. 2 and 3, unlike the fully-supervised approach, $\patern{}^+$ does not require operator preferences over all terrains and is capable of extrapolating to visually novel terrains.

\subsection{Qualitative Large-Scale Experiment}

\begin{figure}[h]
    \centering
    \includegraphics[width=0.484\textwidth]{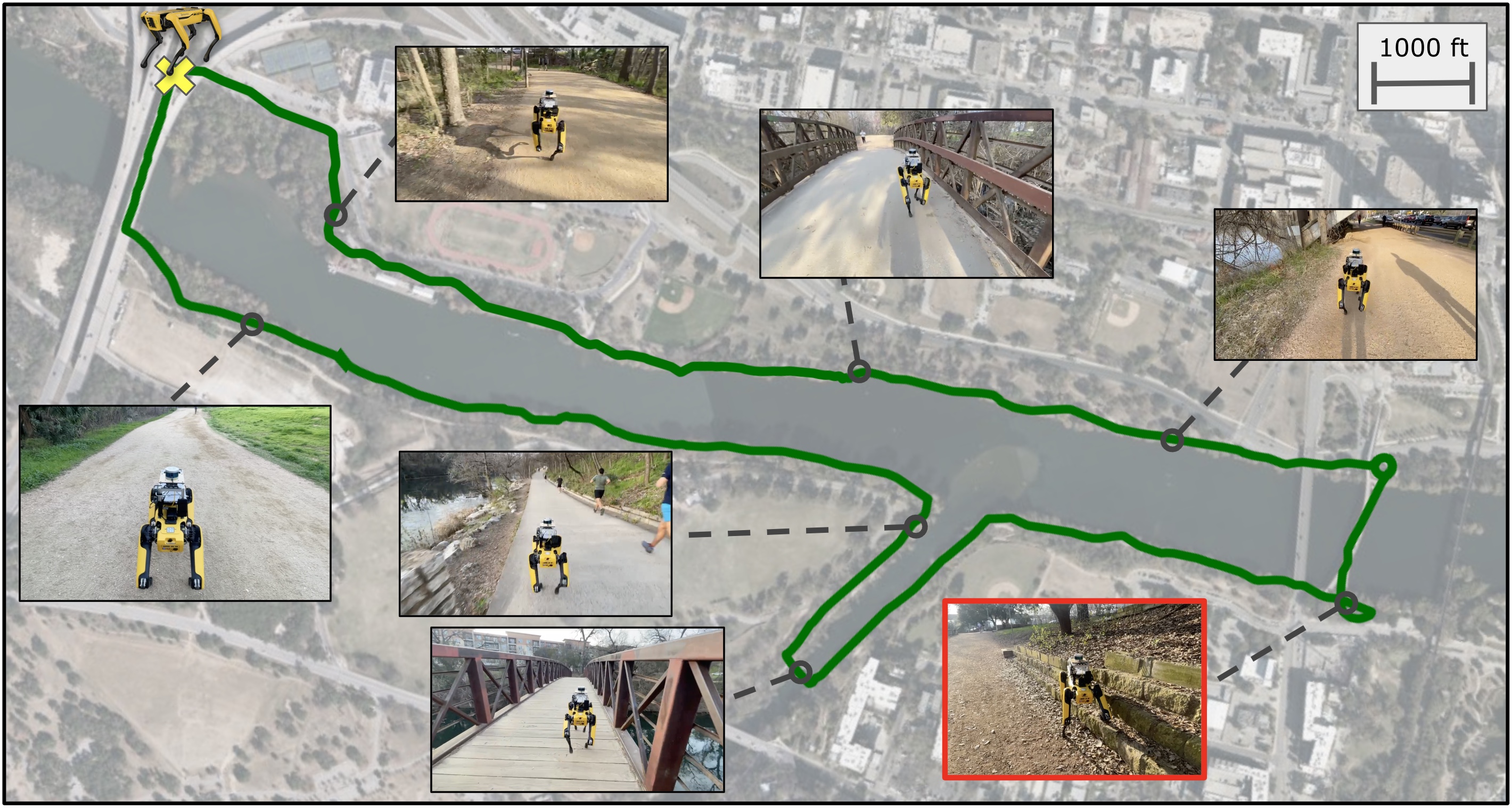}
    \caption{
        Trajectory trace of a large-scale qualitative deployment of $\patern{}^+$ along a 3-mile segment of the Ann and Roy Butler trail located in Austin, Texas. With only five minutes of supplementary data, \patern{} required only one manual intervention to stay on the trail and successfully completes the hike, demonstrating robustness and adaptability to real-world off-road conditions. 
    }
    \label{fig:large_scale}
\end{figure}

To investigate $Q_3$, we execute a large-scale autonomous deployment of \patern{} along a challenging 3-mile off-road trail \footnote{Ann and Roy Butler trail, Austin, TX, USA}. The robot's objective is to navigate in a terrain-aware manner on the trail by preferring \texttt{dirt}, \texttt{gravel} and \texttt{concrete} over \texttt{bush}, \texttt{mulch}, and \texttt{rocks}.
Failure to navigate in a preference-aligned manner may cause catastrophic effects such as falling into the river next to the trail.
An operator is allowed to temporarily take manual control of the robot only to prevent such catastrophic effects, adjust the robot's heading for forks in the trail, or yield to pedestrians and cyclists.
The $\patern{}^-$ model used for short-scale experiments is augmented with approximately five minutes of combined additional data for \texttt{dirt, bush,} and \texttt{mulch} terrains commonly seen in the trail. Following this preference extrapolation, the $\patern{}^+$ model is able to successfully navigate the 3-mile trail, while only requiring one human intervention. Fig. \ref{fig:large_scale} shows the trajectory of the robot and a number of settings along the trail, including the single unexpected terrain-related intervention in the lower right corner, for the hour-long deployment. Additionally, we attach a video recording of the robot deployment\footnote{\href{https://youtu.be/j7159pE0u6s}{\patern{} deployed in the trail: https://youtu.be/j7159pE0u6s}}. This large-scale study addresses $Q_3$ by qualitatively demonstrating the effectiveness of \patern{} in scaling to real-world off-road conditions.

\section{LIMITATIONS AND FUTURE WORK}
\patern{} uses similarities between novel and known terrains in its learned proprioception representation space to extrapolate preferences. Thus, \patern{} needs to have had experiences with terrains bearing close inertial-proprioceptive-tactile resemblances for successful extrapolation. A noticeable limitation is that if a terrain evoking similar proprioceptive features has not been encountered before, \patern{} might not be able to extrapolate preferences. Additionally, \patern{} utilizes non-visual observations that require a robot to physically drive over terrains, which may be unsafe or infeasible in certain cases. Extending \patern{} with depth sensors to handle non-flat terrains is a promising direction for future work.

\section{CONCLUSION}

In this work, we present \textit{Preference extrApolation for Terrain-awarE Robot Navigation} (\patern{}), a novel approach to extrapolate human preferences for novel terrains in visual off-road navigation. \patern{} learns an inertial-proprioceptive-tactile representation space to detect similarities between visually novel terrains and the set of known terrains. Through this self-supervision, \patern{} successfully extrapolates operator preferences for visually novel terrain segments, without requiring additional human feedback. Through extensive physical robot experiments in challenging outdoor environments in varied lighting conditions, we find that \patern{} successfully extrapolates preferences for visually novel terrains and is scalable to real-world off-road conditions.

\newpage
\bibliographystyle{IEEEtran}
\bibliography{mybib}

\end{document}